\documentclass[10pt,twocolumn,letterpaper]{article}

\usepackage{iccv}
\usepackage{times}
\usepackage{epsfig}
\usepackage{graphicx}
\usepackage{amsmath}
\usepackage{amssymb}

\usepackage[sort,nocompress]{cite}
\usepackage{textcomp}
\usepackage[table]{xcolor}
\definecolor{mygray}{gray}{0.95}
\definecolor{Cinnabar}{RGB}{234, 67, 53}
\usepackage{multirow}
\usepackage{times}
\usepackage{epsfig}
\usepackage{graphicx}
\usepackage{amsmath}
\usepackage{amssymb}
\usepackage[boxed]{algorithm}
\usepackage{algorithmic}
\usepackage{soul}
\usepackage{csquotes}
\usepackage{booktabs,subcaption,amsfonts,dcolumn}

\usepackage[pagebackref=true,breaklinks=true,letterpaper=true,colorlinks,bookmarks=false]{hyperref}

\iccvfinalcopy 

\def\iccvPaperID{****} 

\ificcvfinal\fi

\begin{document}

\title{
Social Fabric: Tubelet Compositions for Video Relation Detection}


\author{Shuo Chen, Zenglin Shi, Pascal Mettes, and Cees G. M. Snoek \\
University of Amsterdam
}

\ificcvfinal\thispagestyle{empty}\fi

\twocolumn[{
\renewcommand\twocolumn[1][]{#1}
\maketitle
\begin{center}
    \centering
    \includegraphics[width=0.95\textwidth]{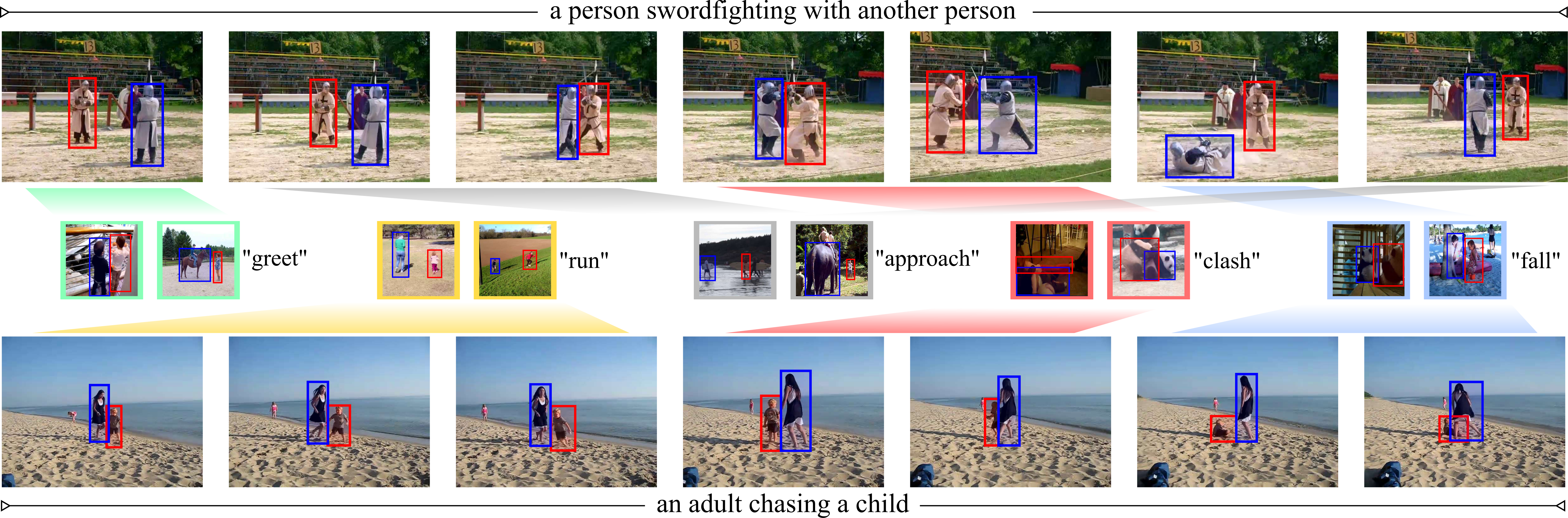}
    \captionof{figure}{\textbf{Social Fabric} encodes compositions of interaction primitives defined over tubelet pairs. The primitives are data driven and may correspond to interactions like ``greet'', ``clash'' and ``fall''. Using the primitives, our two-stage network can classify, detect, and search for complex relations across entire videos.
    }
    \label{fig:fig1}
\end{center}
}]

\begin{abstract}
This paper strives to classify and detect the relationship between object tubelets appearing within a video as a \emph{⟨subject-predicate-object⟩} triplet. Where existing works treat object proposals or tubelets as single entities and model their relations \emph{a posteriori}, we propose to classify and detect predicates for pairs of object tubelets \emph{a priori}. We also propose Social Fabric: an encoding that represents a pair of object tubelets as a composition of interaction primitives. These primitives are learned over all relations, resulting in a compact representation able to localize and classify relations from the pool of co-occurring object tubelets across all timespans in a video. The encoding enables our two-stage network. In the first stage, we train Social Fabric to suggest proposals that are likely interacting. We use the Social Fabric in the second stage to simultaneously fine-tune and predict predicate labels for the tubelets.
Experiments demonstrate the benefit of early video relation modeling, our encoding and the two-stage architecture, leading to a new state-of-the-art on two  benchmarks. We also show how the encoding enables query-by-primitive-example to search for spatio-temporal video relations. Code:  \url{https://github.com/shanshuo/Social-Fabric}.
\end{abstract}

\section{Introduction}
To understand what is happening where in videos, it is necessary to detect and recognize relationships between individual instances. Effectively capturing these relationships could improve  captioning~\cite{show}, video retrieval~\cite{snoek2009concept}, visual question answering~\cite{antol2015vqa} and many other visual-language tasks.
In this paper, we strive to classify and detect the relationship between object tubelets appearing throughout a video as a \emph{⟨subject-predicate-object⟩} triplet, like \emph{⟨dog-chase-child⟩} or \emph{⟨horse-stand\_behind-person⟩}.

Shang \etal~\cite{shang2017video,shang2019annotating} pioneered this challenging problem by their definition of video datasets with dense bounding box annotations, temporal bounds, and relationship-triplet labels. Following their guidance, a leading approach to date is to generate proposals for individual objects on short video snippets, encode the proposals, predict a relation and associate the relations over the entire video, \eg \cite{qian2019video, xie2020video, su2020video}. 
To better detect long-term interactions, Liu \etal~\cite{liu2020beyond} forego the need for snippets by first localizing individual object tubelets throughout the entire video, filter out unlikely pairs and predict predicates for the remaining ones. 
Different from all these existing works on video relation prediction, which treat object proposals or tubelets as single entities and model their relations \emph{a posteriori}, we propose to classify and detect predicates for pairs of object tublets \emph{a priori}. 

Considering objects as tubelet pairs from the start requires an encoding that enables us to localize and classify interactions from the pool of all co-occurring object tubelets across all timespans in a video. This is reminiscent of many classical problems in computer vision that need to aggregate spatial, \eg \cite{sivic2003video, gemertPAMI10, jegou2010aggregating, arandjelovic2016netvlad}, temporal, \eg \cite{wang2016temporal,lin2018nextvlad,yue2015beyond} or spatio-temporal, \eg \cite{girdhar2017actionvlad,miech2017learnable,girdhar2019video} primitives into a common representation.
We take inspiration from ActionVLAD by Girdhar \etal~\cite{girdhar2017actionvlad}, which encodes actions as a composition of local action primitives to capture the entire spatio-temporal extent of actions. In this paper, we also learn to encode local spatio-temporal video features in a compositional manner. Different from ActionVLAD, which operates on an entire video, our Social Fabric encoding operates on tubelet pairs, \ie
on inputs from multiple object tubelets and multiple modalities, with a set of interaction primitives that is dynamically learned during video relation training. Social Fabric captures information across the entire scope of tubelet pairs, which is especially beneficial when interactions last long. See Figure~\ref{fig:fig1} for an illustrative example.

We make three contributions. First, we propose to classify and detect video relations for pairs of object tubelets from the start. Second, we introduce Social Fabric, a compositional encoding suited for multi-tubelet and multi-modal inputs. The interaction primitives that form the encoding are learned and updated dynamically, akin to the NetVLAD layer from Arandjelovi\'c \etal \cite{arandjelovic2016netvlad} for visual place recognition. Third, to leverage the Social Fabric, we propose a two-stage network for video relation classification and detection. In the first stage, we localize interactions by training Social Fabric to propose tubelet pairs that are likely interacting. In the second stage we use the Social Fabric to simultaneously fine-tune and learn to predict predicate labels for the tubelets.
Experiments on the benchmarks for video relation detection of Shang \etal~\cite{shang2017video,shang2019annotating} show the benefits of our approach, especially when interactions are long and complex. Social Fabric outperforms alternative video encodings and our two-stage architecture sets a new state-of-the-art for both video relation classification and detection.
Besides classification and detection, we show that our encoding enables searching for relations in videos by providing primitive-examples as queries.

\section{Related Work}
\textbf{Image relation detection.}
Visual relation recognition has a long-standing tradition for static images~\cite{gupta2009observing, yao2010modeling, chao2015hico, li2019transferable, liao2020ppdm, wan2019pose,Hu_2020_CVPR,Graber_2020_CVPR,Mi_2020_CVPR,Inayoshi_eccv_2020}. Besides recognizing visual relationships between objects, Chao \etal~\cite{chao2018learning} introduce the problem of detecting human-object-interactions in static images and contribute a corresponding dataset. It inspired many to contribute to human-object-interaction detection, \eg~\cite{li2019transferable,wan2019pose,wang2020learning, xu2019learning,Cui_2020_CVPR}. Li \etal~\cite{li2019transferable}, for example, learn the knowledge between human and object categories from the provided datasets and use this knowledge as a prior while performing detection. Wan \etal~\cite{wan2019pose} introduce a pose-aware network that employs a multi-level feature strategy. Where image-based relation detection requires two boxes (subject and object) and a predicate, we aim to perform video-based relation detection, which requires us to also localize and track subjects and objects over time.

\textbf{Snippet relation detection.}
Many before us have investigated relation detection in videos~\cite{shang2017video,tsai2019video,qian2019video,di2019multiple,sun2019video,zheng2019relation,shang2019annotating,liu2020beyond, cao2021relation,xie2020video,su2020video,Kukleva_2020_CVPR,Sunkesula_mm_2020}. Relation in videos provide additional temporal information, important for interactions such as pushing or pulling a closed door. Shang \etal~\cite{shang2017video} pioneered this problem and introduced the ImageNet-VidVRD dataset, the first video relation detection benchmark in which all video relation triplets, along with their object and subject trajectories, are labelled. Building on the foundational work of Shang \etal~\cite{shang2017video}, Tsai \etal~\cite{tsai2019video} propose a gated spatio-temporal energy graph using conditional random fields to model video relations. In a similar spirit, Qian \etal~\cite{qian2019video} built a spatio-temporal graph between adjacent video snippets and used multiple layers of graph convolutional networks to pass messages between nodes. Shang \etal~\cite{shang2019annotating} later introduced VidOR, the largest video relation detection benchmark to date. On this dataset, Sun \etal~\cite{sun2019video} utilize language context features along with spatio-temporal features for predicate prediction.

All the aforementioned methods adopt a three-stage framework. A video is first split into short snippets and subject/object tubelets are generated per snippet. Then, short-term relations are predicted for each tubelet. The subject/object proposals are obtained in the short snippets using an image object detector and tracker~\cite{shang2017video, tsai2019video, qian2019video}.
In the second stage, spatio-temporal features of each pair of object tubelets are extracted and used to predict short-term relation candidates. Xie~\etal~\cite{xie2020video} combine a wide variety of multi-modal features for each pair to predict the relations with impressive relation classification accuracy. In the third stage, the short-term relation proposals are merged by a greedy relational association algorithm. Su~\etal~\cite{su2020video} maintain multiple relation hypotheses during the association process to accommodate for inaccurate or missing proposals in the earlier steps. Instead of treating the relations independently at the various analysis stages, we consider the objects tubelets as interacting pairs from the start.

\textbf{Proposal relation detection.}
Liu \etal~\cite{liu2020beyond} are the first to avoid the need to split videos into snippets. In a first stage they generate object tubelets for the whole videos. The second stage refines the tubelet-features and finds relevant object pairs using a graph convolutional network. The third stage focuses on predicting the predicates between related pairs. In this manner, interactions can be detected without a need for snippet splitting. Like Liu \etal, we also avoid the need for snippets.
Different from them, we view subjects and objects as interactions from the start. As a result, we only need two stages, one for interaction proposal generation from the tubelet pairs and one for predicting the appropriate predicate. At the core of both our stages is the Social Fabric, which allows us to encode a set of interaction primitives, like the ones in Figure~\ref{fig:fig1}, from which we classify and detect different video relations. 

\section{Social Fabric Encoding}
The goal in video relation detection is to localize interactions between two entities in space and time. Formally, a spatio-temporal interaction $\mathcal{I}$ is defined as a triplet $\mathcal{I} = \{O_1,P,O_2\}$, with subject tubelet $O_1 \in \mathbb{R}^{4 \times (T_2 - T_1)}$, object tubelet $O_2 \in \mathbb{R}^{4 \times (T_2 - T_1)}$ and their relation predicate category $P$. Here, $T_1$ and $T_2$ denote the start and end frame of the interaction and each frame contains box coordinates. To address both video relation classification and detection, we propose a two-stage approach that encodes subjects and objects as pairs from the start. Central to both stages is our Social Fabric encoding for representing compositions of tubelet pairs. Below, we outline how to learn the encoding, how to use it to represent tubelet pairs and how the encoding relates to existing video encodings.

\begin{figure}[t!]
\centering
    \includegraphics[width=.95\linewidth]{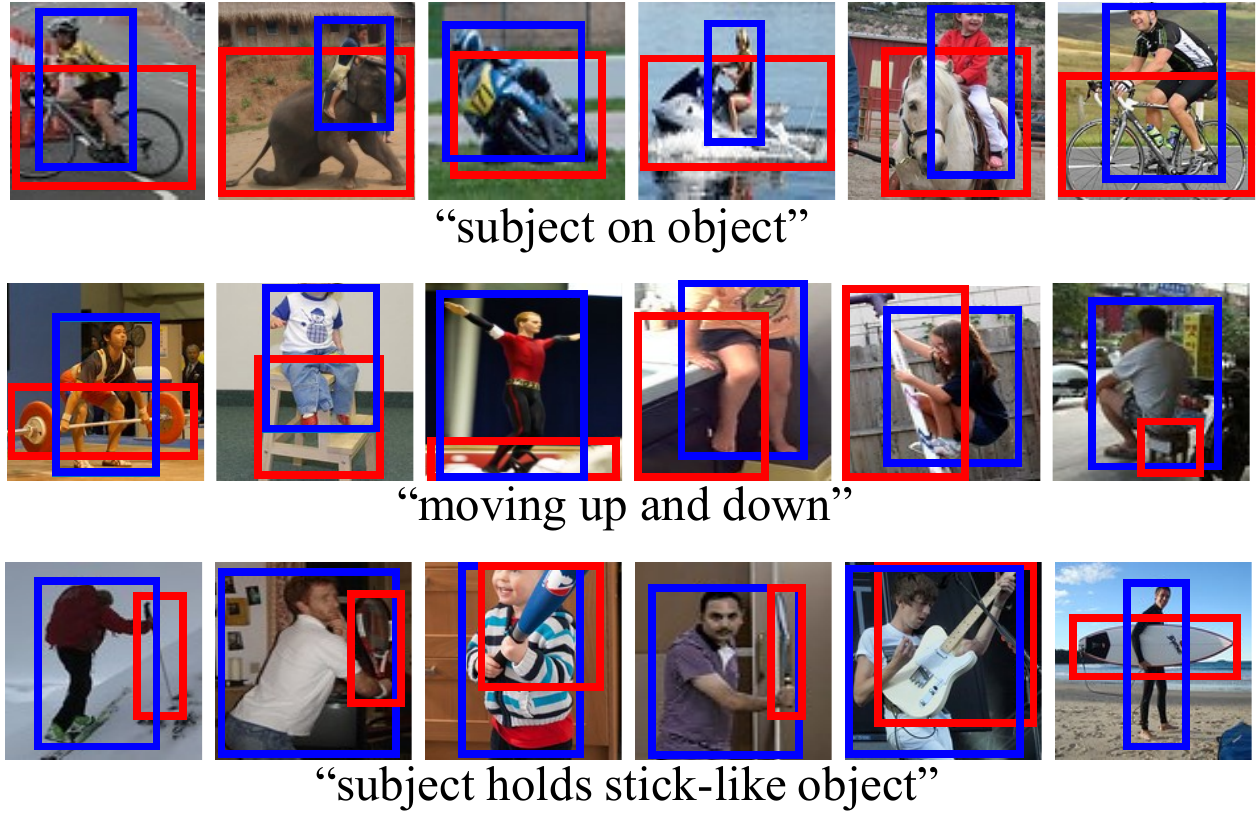}
  \caption{\textbf{Interaction primitives} that our Social Fabric encoding learns when trained for multi-modal features. Each row shows several frames from videos that get assigned to one specific primitive. Blue boxes indicate the subject while red boxes denote the object. Here we show some easy to interpret primitives. 
  }
\label{fig:primitives}
\end{figure}

\textbf{Learning the encoding.}
The idea behind the encoding is that a pair of tubelets, which form a video relation triplet, are composed of multiple interaction primitives. These primitives can represent different relations by varying their combinations. For example, let \{``approach",``run",``watch",``touch"\} denote a set of primitives, then a hugging relation can be represented by \{``watch",``approach",``touch"\}, while a chasing relation can be represented by \{``run",``approach"\}. In the object detection and action recognition literature, compositional learning and encoding is well established, with advantages such as sharing components amongst categories \eg \cite{gaidon2013temporal}, efficient and compact encoding \eg \cite{yuille2011towards}, and high discriminative ability \eg \cite{juneja2013blocks, kortylewski2020compositional}. By introducing a compositional encoding for video relation detection we share the same benefits and show some examples of the primitives we learned in Figure~\ref{fig:primitives}.

\begin{figure}[b!]
\centering
    \includegraphics[width=.85\linewidth]{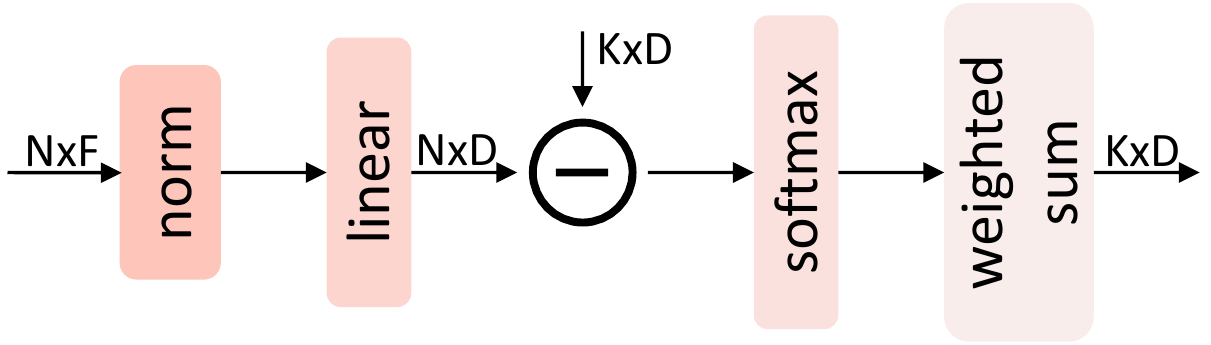}
    \vspace{-0.1cm}
  \caption{\textbf{Social Fabric Encoding}.
  }
\label{fig:SFE}
\end{figure}

For each task, we are given a training set of tubelet pairs, denoted as $\mathcal{R}$, where the input representation of each tubelet pair is denoted as $S_i \subset \mathcal{R} \in \mathbb{R}^{N \times F}$, with $N$ the number of frames of the tubelets and $F$ the feature dimensionality for each frame, denoting the combined subject and object representations. On top of the features, we apply layer normalization~\cite{ba2016layer}, followed by a linear layer to obtain embedded representation $R_i \subset \mathcal{R} \in \mathbb{R}^{N \times D}$. In this $D$-dimensional embedding space, we learn a set $C \in \mathbb{R}^{K \times D}$ consisting of $K$ primitives. The idea behind our encoding is to describe a tubelet pair entirely as a weighted combination of these primitives. So tubelet pair $i$ is encoded with our approach as a concatenation of weighted primitive locations:
\begin{equation}
E_i {=} [E_{i,1},\cdots,E_{i,K}], \
E_{i,k} {=} \sum_{j=1}^N {z_{ijk} C_{k}},
\end{equation}
where the weight is inversely proportional to the distance between a local relational feature vector and the primitive:
\begin{equation}
z_{ijk} = \frac{\exp \left[ -\beta  \left\| R_{ij} - C_k \right\|^2 \right]}{\sum_{l=1}^{K} \exp \left[ -\beta \left\| R_{ij} - C_l \right\|^2 \right]}, \label{equ:zij}
\end{equation}
where $\beta > 0$ denotes a temperature parameter to tune how soft or hard the assignments should be, fixed to $1/\sqrt{D}$ throughout this work. Intuitively, our encoding describes how much a relation is in line with each primitive in $C$. Each portion $E_{i,k}$ of the encoding forms a line between the primitive $C_k$ and the origin; the stronger the agreement, the closer $E_{i,k}$ is to the primitive and the more its values contribute to the next layer. The diagram of the Social Fabric Encoding is shown in Figure~\ref{fig:SFE}.

\begin{figure*}[t]
\begin{center}
\includegraphics[width=.95\linewidth]{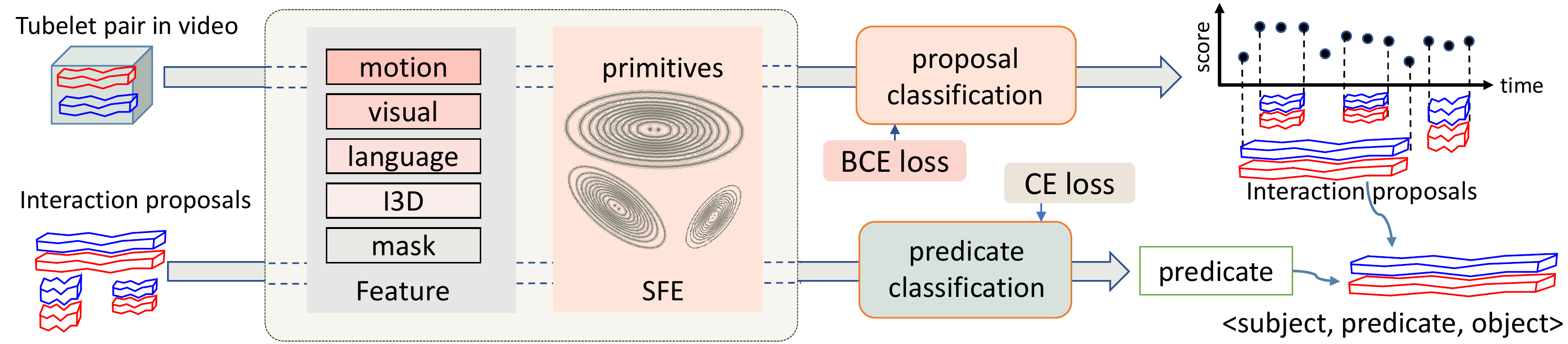}
\end{center}
  \caption{
\textbf{Two-stage video relation network}. We first obtain interaction proposals and then predicate predictions. Social Fabric Encoding (SFE) is essential to both stages as to represent an object tubelet with a composition of interaction primitives. BCE loss and CE loss represents binary cross-entropy loss and cross-entropy loss separately.}
\label{fig:overview}
\end{figure*}

On top of the representation $E_i$, we learn a fully-connected layer classification head, which can be used to determine whether a tubelet pair makes for a good proposal or to predict its predicate using a shallow network head. The layers of the network and the set $C$ are jointly learned during the optimization. 

\textbf{Relation to alternative encodings.}
A common encoding in video-based representations is average pooling~\cite{yue2015beyond}. In our encoding, average pooling is a special case where the codebook contains a single primitive. Average pooling implicitly assumes that the features of the input representation follow a single mode. Video relations, however, consist of multiple interaction primitives that evolve over time. Moreover, these primitives are shared between different relations, which we capture.
Encodings such as transformers follow the self-attention architecture, where each feature is a weighted sum of other features~\cite{vaswani2017attention}. Compared to transformers, our approach provides a fixed-sized representation, important because tubelet pairs are of varying length. Other encodings like NetVLAD~\cite{arandjelovic2016netvlad} and ActionVLAD~\cite{girdhar2017actionvlad}   operate on whole images and videos, while residuals between local features and clusters are used to obtain a representation. In contrast, our encoding operates on pairs of spatio-temporal tubelets, accepts multi-modal features, and we directly use the primitives to encode inputs. Lastly, we are the first to rely on a compositional encoding for the task of video relation detection.

\section{Two-stage video relation network}
We utilize the Social Fabric Encoding to both classify and detect video relations using two stages, rather than three stages common in the literature. In the first stage, we sift through all combinations of co-occurring tubelets across all timesteps to obtain a set of interaction proposals that likely cover all ground truth video relations. In the second stage, we classify each proposal with a predicate label. An overview of our approach is visualized in Figure~\ref{fig:overview}. Next, we detail both stages and show how to obtain the final classification and spatio-temporal detection results.

\textbf{Stage 1: Interaction proposals.}
We initialize the video relation optimization by performing object detection in each frame, followed by linking over time based on~\cite{2017ICIP-DeepSort}. For a video $V$, this results in $M$ object tubelets. We consider all unique combinations of tubelets for proposal generation and train a binary classifier to determine interactivityness at the frame-level using a local window around the box pairs in a frame \cite{chen2020interactivity}.
For the two objects $(O_1,O_2)$ in a tubelet pair and frame $f$, we consider a neighbourhood of $m/2-1$ frames in both temporal directions of the tubelets. We compute and stack the multi-modal features for the windowed tubelet pair, resulting in $R^1_f(O_1,O_2) \in \mathbb{R}^{m \times D}$ for frame $f$. We feed this as input to Social Fabric, resulting in $E^1(O_1,O_2) \in \mathbb{R}^{K \times D}$. During training, the encoding is used to train a binary classifier to separate potential interactions from non-interactions with a binary cross-entropy loss $\mathcal{L} {=} \big(y \log(s) + (1-y) \log(1-s) \big)$, where $s$ denotes the interactivityness.
 Simultaneously, the primitives in the Social Fabric are learned. For each frame in a tubelet pair, this results in a score indicating its interactivityness. Over the array of scores over all timesteps of the tubelet pair, we employ a 1D watershed algorithm~\cite{chen2020interactivity,2000-watershed} to generate spatio-temporal interaction proposals. We repeat this procedure for all co-occurring tubelets and combine the outputs per pair into a final set of interaction proposals for a video.

\textbf{Stage 2: Predicate prediction.}
Once a video is decomposed into a set of interaction proposals, each consisting of two tubelets with a similar start and end time, we seek to score all proposals for their predicate.
For interaction proposal $(O_1, O_2)$, we sample $n$ frames uniformly.
For each sampled frame, we extract a single uni-modal or several multi-modal features. Then we stack the features over all frames and obtain $R^2(O_1, O_2) \in \mathbb{R}^{N \times D}$ for this tubelet. This is fed into Social Fabric and the output representation is in $E^2(O_1,O_2) \in \mathbb{R}^{K \times D}$. 
In stage 2 we fine-tune the Social Fabric trained in stage 1 to accelerate the convergence. After encoding each proposal, we feed the representation into a final linear layer to obtain predicate scores. The predicate prediction is optimized with softmax cross-entropy.
After obtaining predicate predictions, we multiply the predicate score and corresponding subject and object scores as the relation triplet prediction score. The subject and object scores are obtained from the tubelet pairs in stage 1. Relation triplets are the predicted results for relation classification. The relation triplet associated with subject and object tubelets act as the predicted results for relation detection.

\textbf{Search-by-primitive-example.} The Social Fabric encoding is optimized for video relation classification and detection, but is not limited to these tasks. Here, we show how we can also search for spatio-temporal video relations in a collection of videos by querying primitive examples. 
As input, a user can provide one or more frames with a subject and object performing a basic interaction. We compute the non-temporal features for each input and use it to find the nearest learned primitive. To find the interaction proposal across all videos that best describes the primitive examples, we use the weights from Equation~\ref{equ:zij} to score the relevance of each primitive for an entire proposal. In turn, we simply sum the scores for the few primitives determined by the user and output the interaction proposal with the highest score. As a result, we can search on-the-fly for video relations that are composed of  example primitives provided by a user, without the need for search optimization.

\begin{table*}[t]
\centering
\resizebox{0.8\linewidth}{!}{
\begin{tabular}{cccccrrrrrr}
\toprule
\multicolumn{5}{c}{Feature type} 
& \multicolumn{3}{c}{Relation tagging} & \multicolumn{3}{c}{Relation detection} \\
\cmidrule(lr){1-5} \cmidrule(lr){6-8} \cmidrule(lr){9-11}
motion & visual & language & I3D & mask & P@1 & P@5 & P@10 & mAP & R@50 & R@100  \\
\midrule
\checkmark & & & && 50.97 & 39.57 & 31.58 & 6.14 & 6.74 & 8.70  \\
\checkmark &\checkmark & & && 56.89 & 44.76 & 34.07 & 8.93 & 7.38 & 9.22  \\
\checkmark &\checkmark &\checkmark & && 59.24 & 47.24 & 35.99 & 9.54 & 8.49 & 10.17  \\
\checkmark &\checkmark &\checkmark &\checkmark && 61.52 & 50.05 & 38.48 & 10.04 & 8.94 & 10.69  \\
\rowcolor{mygray} \checkmark &\checkmark &\checkmark &\checkmark &\checkmark & \textbf{68.86} & \textbf{55.16} & \textbf{43.40} & \textbf{11.21} & \textbf{9.99} & \textbf{11.94}  \\
\bottomrule
\end{tabular}
}
\caption{\textbf{Benefit of multi-modal features} on VidOR. More is better. The increasing gaps indicate Social Fabric effectively captures multi-modal features for relation classification and detection.}
\label{tab:type}
\end{table*}

\section{Experimental setup}

\subsection{Datasets}
To evaluate the proposed methods, we perform experiments on ImageNet-VidVRD \cite{shang2017video} and Video Object Relation (VidOR) \cite{shang2019annotating}.

\textbf{ImageNet-VidVRD.} \cite{shang2017video} consists of 1,000 videos, created from the ILSVRC2016-VID dataset \cite{russakovsky2015imagenet}. There are 35 object categories and 132 predicate categories. The videos are densely annotated with relation triplets in the form of \emph{⟨subject-predicate-object⟩} as well as the corresponding subjects and objects trajectories. Following \cite{shang2017video, tsai2019video}, we use 800 videos for training and the remaining 200 for testing. 

\textbf{VidOR.} \cite{shang2019annotating} contains 10,000 user-generated videos selected from YFCC-100M \cite{thomee2016yfcc100m}, for a total of about 84 hours. There are 80 object categories and 50 predicate categories. Besides providing annotated relation triplets, the dataset also provides the bounding boxes of objects. The dataset is split into a training set with 7,000 videos, validation set with 835 videos, and a testing set with 2,165 videos. Since the ground truth of the test set is not available, we use the training set for training and the validation set for testing, following \cite{liu2020beyond,qian2019video, xie2020video, su2020video}.

\subsection{Implementation and evaluation details}
\label{sec:implementation}


\textbf{Tubelet pairing.} We first detect all the objects per video frame by Faster R-CNN~\cite{2015NIPS-faster} with a ResNet-101~\cite{2016CVPR-ResNet} backbone. The detector is trained on MS-COCO~\cite{lin2014microsoft}. The detected bounding boxes are linked with the Deep SORT tracker~\cite{2017ICIP-DeepSort} to obtain individual object tubelets. 
Finally, each tubelet is paired with any other tubelet to generate the tubelet pairs. We use the object trajectories of ImageNet-VidVRD and VidOR adopted in \cite{shang2017video, qian2019video, su2020video, sun2019video} for fair comparison. 

\textbf{Feature extraction.}
In the video relation literature, features from multiple modalities are commonly used, \eg
Sun \etal \cite{sun2019video} use motion features and language features. Liu \etal \cite{liu2020beyond} use motion features, visual features and I3D features. Xie \etal \cite{xie2020video} use motion features, visual features, language features and location mask features. 
We consider all features and arrive at motion features, visual features, language features, I3D features, and location mask features. 
We follow \cite{sun2019video} to calculate the spatial location feature as motion features. The visual features are extracted using the detection backbone in Faster R-CNN and followed by an RoI pooling layer. For the language features we use a word2vec module, pre-trained on GoogleNews~\cite{mikolov2013efficient}, to encode the subject and object classes into language features with dimension of 600. We use the I3D module from \cite{carreira2017quo} to extract I3D features with fixed dimension of 832. We follow the method of \cite{xie2020video} to generate a mask based on the bounding boxes of the subject and object in the tubelet pair.

\textbf{Two-stage network optimization.} The size of the linear layer for embedding representation is $D{=}512$. In the first stage, we consider $m{=}30$ neighbourhood frames on both temporal directions.
The interaction proposal generation network is trained for 20 epochs using an SGD optimizer with a mini-batch of 128. We use a fixed learning rate and set its value to 0.01. 
In the second stage, we sample $n{=}25$ frames for each interaction proposal. The predicate prediction network is trained for 10 epochs using an SGD optimizer with a mini-batch of 128. We use a fixed learning rate and set its value to 0.01. 

\textbf{Evaluation metrics.} Following \cite{shang2017video},
we adopt Precision@$1$, Precision@$5$ and Precision@$10$ to measure the ability of classifying visual relations. We will refer to the classification task as relation tagging in the experiments for consistency with current literature. For video relation detection we report mAP (mean Average Precision), Recall@$50$ and Recall@$100$.

\section{Results}

\textbf{Benefit of multi-modal features.}
We first evaluate the benefit of the use of multi-modal features on VidOR in Table~\ref{tab:type}. With only motion features, our method achieves a P@1 of 50.97 for relation tagging and an mAP of 6.14 for relation detection. With all features included, the performance is clearly improved with a P@1 of 68.86 for relation tagging and an mAP of 11.21 for relation detection.
The results show that our encoding benefits from incorporating information from many modalities. In the following ablations, we use all features.

\textbf{Influence of encoding size.} 
Next, we evaluate the influence of the number of interaction primitives in the Social Fabric Encoding. Intuitively, the more primitives, the finer commonalities between interactions are modelled. In Table~\ref{tab:iter}, we find that multiple primitive components indeed improves over a single component (which resembles conventional average pooling).
When increasing the number of primitives, we further improve the performance. The Social Fabric Encoding performs best at $K{=}64$, where it provides a balance between coverage of the space and sharing amongst relations. We use this encoding size for further experiments.

\begin{table}[t]
\centering
\begin{tabular}{cccccc}
\toprule
 Clusters  & 1 & 8 & 32 & 64 & 128 \\
\midrule
mAP & 10.05 & 10.69 & 10.91 & \textbf{11.21} & 11.01 \\
\bottomrule
\end{tabular}
\caption{\textbf{Influence of encoding size} on VidOR for relation detection. Using multiple primitives results in a more accurate predicate prediction, where we achieve best performance for 64 primitives.}
\label{tab:iter}
\end{table}

\textbf{Importance of two stages.}
Next, we show the importance of the interaction proposal stage and the predicate predication stage on VidOR in Table~\ref{tab:stage}. 
The baseline (first row) splits the video into short snippets. Relationships are separately detected in each snippet and merged afterwards, akin to \cite{qian2019video, xie2020video, su2020video}. It average pools the features before predicate prediction. With the interaction proposal stage added (second row), we have spatio-temporal proposals covering long-range interactions. It provides the necessary context to recognize long duration interactions. Accordingly, both recall and precision improve. The Recall@50 is improved by 1.09 and P@1 is improved by 3.47 compared to the baseline. Upon adding the second stage (Third row), the P@1 increases by 4.67 compared to when we only use interaction encoding in proposal generation. We conclude that both stages matter in combination with our encoding.

\begin{table}[t!]
\centering
\resizebox{\linewidth}{!}{
\begin{tabular}{cccccccc}
\toprule
 &  & \multicolumn{3}{c}{Relation tagging} & \multicolumn{3}{c}{Relation detection}\\
\cmidrule(lr){3-5} \cmidrule(lr){6-8}
Stage 1 & Stage 2 & P@1 & P@5 & P@10    & mAP & R@50 & R@100\\
\midrule
&  &	60.72 &	46.40 &	36.62   & 9.61 &	8.73 &	10.81\\
\checkmark &  & 64.19 & 49.60 & 39.22  & 10.16 & 9.62 & 11.63\\
\rowcolor{mygray} \checkmark & \checkmark  & \textbf{68.86} & \textbf{55.16} & \textbf{43.40}  & \textbf{11.21} & \textbf{9.99} & \textbf{11.94}\\
\bottomrule
\end{tabular}
}
\caption{\textbf{Importance of two stages} on VidOR. Incorporating Social Fabric into the two stages of our pipeline (third row) is preferred over baselines based on average pooling of features with video snippet proposals (first row) and using Social Fabric only for the proposals (second row).}
\label{tab:stage}
\end{table}

\begin{table}[t]
\centering
\begin{tabular}{lcc}
\toprule
& Relation tagging & Relation detection \\
\cmidrule(lr){2-2} \cmidrule(lr){3-3}
Encoding & P@1 & mAP \\
\midrule
average pooling & 62.73 & 10.05 \\
transformer & 63.86 & 10.07 \\
NetVLAD & 65.34 & 10.15 \\
NetRVLAD & 66.80 & 10.55 \\
\rowcolor{mygray} Social Fabric & \textbf{68.86} & \textbf{11.21} \\
\bottomrule
\end{tabular}
\caption{\textbf{Comparison with alternative encodings} on VidOR. Social Fabric performs well.
}
\label{tab:encodingcompare}
\end{table}

\begin{table*}[t!]
\centering
\resizebox{1\linewidth}{!}{
\begin{tabular}{lrrrrrrrrrrr}
\toprule
& \multicolumn{6}{c}{\textbf{ImageNet-VidVRD}} & \multicolumn{5}{c}{\textbf{VidOR}}\\
\cmidrule(lr){2-7} \cmidrule(lr){8-12}
& \multicolumn{3}{c}{Relation tagging} & \multicolumn{3}{c}{Relation detection} & \multicolumn{2}{c}{Relation tagging} & \multicolumn{3}{c}{Relation detection}\\
\cmidrule(lr){2-4} \cmidrule(lr){5-7} \cmidrule(lr){8-9} \cmidrule(lr){10-12}
  & P@1 & P@5 & P@10 & mAP & R@50 & R@100 & P@1 & P@5 & mAP & R@50 & R@100 \\
\midrule
 Shang \etal~\cite{shang2017video}   & 43.00 & 28.90 & 20.80 & 8.58 & 5.54 & 6.37 & - & - & - & - & -\\ 
 Tsai \etal~\cite{tsai2019video}          & 51.50 & 39.50 & 28.23 & 9.52 & 7.05 & 8.67  & - & - & - & - & -\\
 Qian \etal~\cite{qian2019video}        & 57.50 & 41.00 & 28.50 & 16.26 & 8.07 & 9.33  & - & - & - & - & -\\
 Sun \etal \cite{sun2019video} & - & - & - & - & - & - &  51.20 & 40.73 & 6.56 & 6.89 & 8.83  \\
 Su \etal~\cite{su2020video}              & 57.50 & 41.40 & 29.45 & 19.03 & 9.53 & 10.38 & 50.72 & 41.56 & 6.59 & 6.35 & 8.05 \\
 Liu \etal~\cite{liu2020beyond}          & 60.00 & 43.10 & 32.24 & 18.38 & 11.21 & 13.69 & 48.92 & 36.78 & 6.85 & 8.21 & 9.90  \\
 Xie \etal~\cite{xie2020video} & - & - & - & - & - & - & 67.43 & - & 9.93 & 9.12 & - \\ 
\midrule
 \textit{\textbf{This paper}}, features as Su \etal~\cite{su2020video} & 57.50 & 43.40 & 31.90 & 19.23 & 12.74 & 16.19 & 54.57 & 43.58 & 8.93 & 9.15 & 11.13 \\
 \textit{\textbf{This paper}}, features as Liu~\etal~\cite{liu2020beyond} & 61.00 & 47.50 & 36.60 & 19.77 & 12.91 & 16.32  & 55.40 & 45.74 & 9.13 & 9.36 & 11.30 \\
 \textit{\textbf{This paper}}, features as Xie~\etal~\cite{xie2020video} & - & - & - & - & - & - &  68.62 & 53.34 & 11.05 & 9.91 & 11.89 \\
\rowcolor{mygray} \textit{\textbf{This paper}}, our features  & \textbf{62.50} & \textbf{49.20} & \textbf{38.45} & \textbf{20.08} & \textbf{13.73} & \textbf{16.88} & \textbf{68.86} & \textbf{55.16} & \textbf{11.21} & \textbf{9.99} & \textbf{11.94} \\
\bottomrule
\end{tabular}
}
\caption{\textbf{Comparison with state-of-the-art} for relation tagging and detection on ImageNet-VidVRD and VidOR. We outperform the recent snippet relation detection methods of both Su \etal and Xie \etal for almost all metrics when using their features. We also outperform the proposal relation detection method of Liu \etal when using their features. When we rely on our full set of features results improve further and set a new state-of-the-art on both tasks for both benchmarks.
}
\label{tab:sota}
\end{table*}

\textbf{Comparison with alternative encodings.}
We compare to the following encodings on VidOR: average pooling, transformer encoding, NetVLAD~\cite{girdhar2017actionvlad}, NetRVLAD~\cite{miech2017learnable}. 
Average pooling corresponds to our encoding with a single mixture component. Transformers were proposed in~\cite{vaswani2017attention} for textual sequence-to-sequence tasks and recently adopted in  video tasks~\cite{byvshev2020heterogeneous,girdhar2019video,gavrilyuk2020actor}. 
Here, we investigate their potential for interaction detection. We feed the frame-level representations to the transformer encoder. The output representation is average pooled and then fed into the predicate classifier. NetVLAD was first introduced for place recognition and later adopted for video action classification in~\cite{girdhar2017actionvlad}. We train a classifier over the NetVLAD layer initialized by $k$-means on all features to initialize the cluster centroids (and keep it fixed). As our method, we use 64 cluster centroids. NetRVLAD~\cite{miech2017learnable} is a simplification of the original NetVLAD architecture that averages the actual descriptors instead of the residuals.

We report the P@1 and mAP on VidOR dataset in Table~\ref{tab:encodingcompare}. 
All encodings take the same multi-modal representations as input. 
The transformer and average pooling baselines obtain similar performance. NetVLAD improves over average pooling and transformers, highlighting the effectiveness of codebook-based encodings. NetRVLAD further improves over NetVLAD, potentially because aggregating the actual feature instead of residuals may benefit the performance~\cite{douze2013stable}. Our encoding uses a similar strategy with a dynamic learning scheme and outperforms all baselines, with an mAP of 11.21\% compared to 10.55\% for NetRVLAD as the best performing alternative.

\textbf{Comparison with state-of-the-art.} We compare with the state-of-the-art in video relation classification and  detection in Table~\ref{tab:sota} for both ImageNet-VidVRD and VidOR. Liu \etal \cite{liu2020beyond} report good results for relation classification and detection on both sets. When we compare with them using the same input features, \ie visual, I3D and motion feature, we improve over their work on all metrics. Most notably, the mAP for relation detection improves from 18.38 to 19.77 on ImageNet-VidVRD and from 6.85 to 9.13 on VidOR. We also compare favorably against the recent snippet-based video relation detection of Su \etal~\cite{su2020video} using their features. We are on par for the relation classification P@1 on ImageNet-VidVRD, but outperform them on all other metrics and datasets, demonstrating the benefit of detecting predicates for social tubelets from the start. Xie \etal~\cite{xie2020video} improved the state-of-the-art considerably by combining a motion feature, visual feature, language feature and location mask feature for each trajectory pair before predicting their relation. Our method profits from such a rich set of multi-modal features also. When we use the same features as Xie \etal our results get better as well, obtaining 68.62 P@1 and 11.05 mAP for relation classification and detection respectively.
Our features adds I3D feature to the feature set used by Xie~\etal~\cite{xie2020video}. Using our features we obtain state-of-the-art performance with 11.21 mAP and 68.86 P@1.
We also consider the computational aspects of our method. We test using a GTX 1080 Ti GPU. With the same features as Liu~\etal~\cite{liu2020beyond}, the average time to process one ImageNet-VidVRD validation video is 58.2s for Liu~\etal~\cite{liu2020beyond}, and 48.3s for our method.

\begin{figure}[t!]
    \centering
    \includegraphics[width=\linewidth]{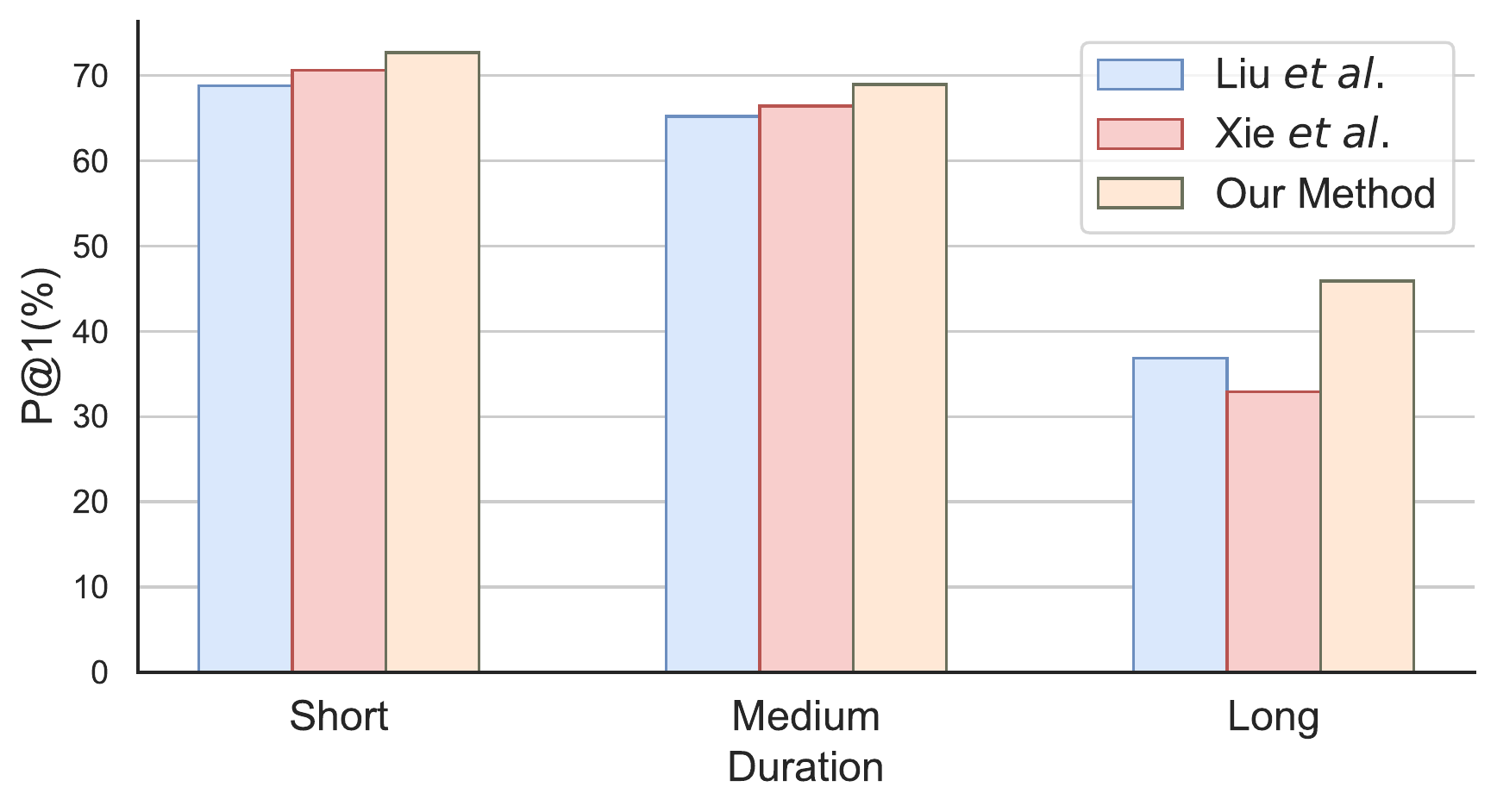}
    \caption{\textbf{Comparison along relation duration} on VidOR. We observe our method's performance improves over alternatives as the duration of the video relation increases. }
\label{fig:length}
\end{figure}


\begin{figure*}[htb!]
\centering
\includegraphics[width=\textwidth]{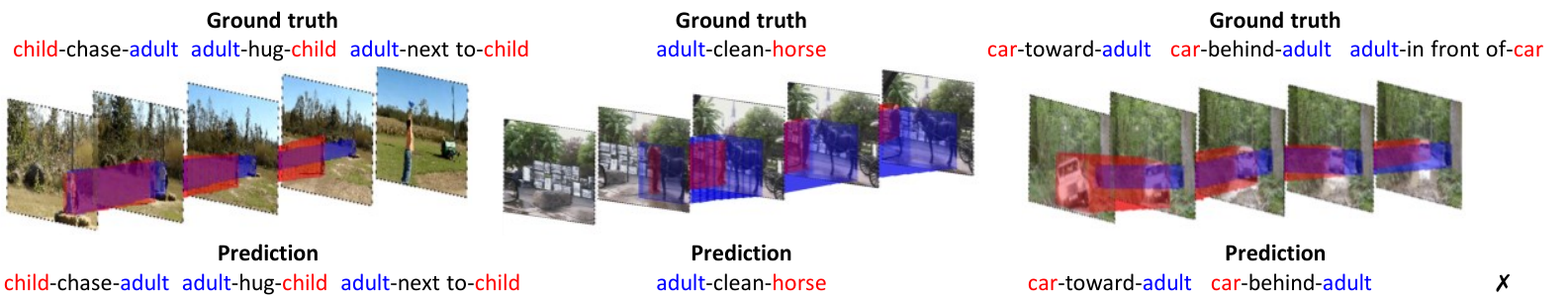}
\vspace{-0.6cm}
  \caption{
\textbf{Success and failure cases} on VidOR. For the left example, we detect all the ground truth relation instances and successfully predict the long-range relation \texttt{chase}. The middle case needs temporal context information to detect an adult cleaning a horse. Our method's detection proves its ability to detect long-range relations. In the right example, our approach detects \texttt{behind} and \texttt{toward} relations. But since the object detector wrongly recognizes \texttt{car} as \texttt{truck}, the final triplet predictions are wrong even though the relation predicates are correct. Incorrect object categories also lead to imprecise semantic features, which may contribute to the missing of a relation prediction. We provide more qualitative results and example videos with success and failure in the supplemental material.
}
\label{fig:cases}
\end{figure*}

\textbf{Comparison along relation duration.} To verify the effectiveness of our approach on long-range relations. we break down the performance into three bins according to the duration of the relation instances: ``short'', ``medium'' and ``long''. We compare our method with Liu~\etal~\cite{liu2020beyond} and Xie~\etal~\cite{xie2020video} on the VidOR validation set. Results are shown in Figure~\ref{fig:length}. The three methods use the same features as Xie~\etal~\cite{xie2020video} for fair comparison. The results of Xie~\etal~\cite{xie2020video} are provided by the authors. The results of Liu~\etal~\cite{liu2020beyond} are obtained by running the provided code.
As expected, Liu~\etal~\cite{liu2020beyond} surpasses Xie~\etal~\cite{xie2020video} for long duration relations as they are designed to be effective beyond short-snippets. Our method is beyond both Liu \etal and Xie \etal for all durations. Compared to Xie~\etal~\cite{xie2020video} who do not consider long-range relations, our method's performance gain increases as the relation length increases. We conclude our approach is beneficial for encoding multi-modal features for relation detection especially at long-range. Besides, we have split the predicates in VidOR into two super categories: action-based and spatial-based relations, following [37]. We obtain a mAP of 7.33\% for action-based relations and a mAP of 12.89\% for spatial-based relations, while the state-of-the-art by Xie 
\etal [51] obtains a mAP of 6.25\% for action-based relations and a mAP of 11.23\% for spatial-based relations.
We show some success and failure cases in Figure~\ref{fig:cases}.

\begin{figure}[t]
\centering
\includegraphics[width=\linewidth]{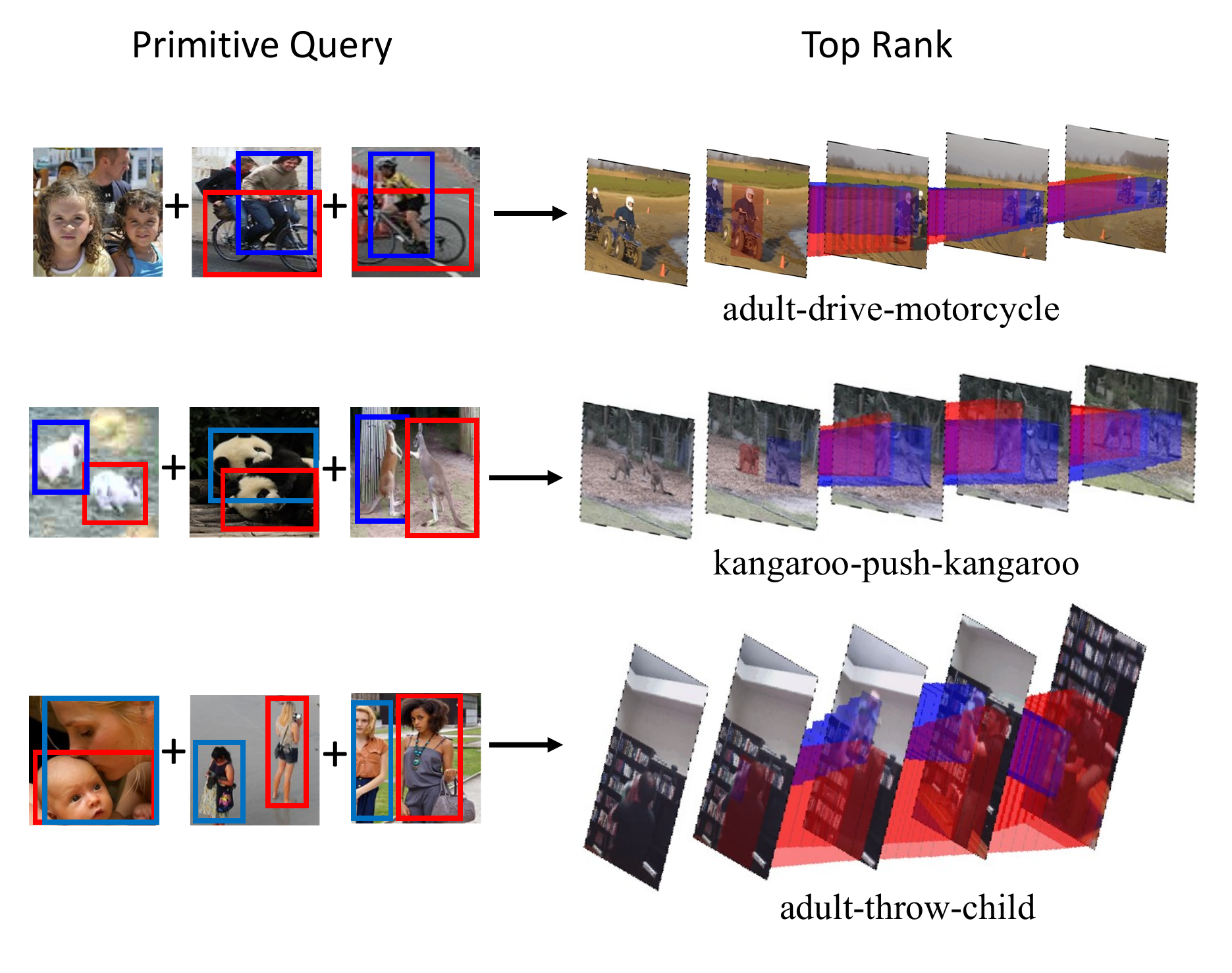}
\vspace{-0.8cm}
\caption{\textbf{Query-by-primitive-examples.} We use three examples of primitives as query. Among the VidOR validation set, the relation whose primitive weights are  closest to the three examples is selected. \eg, in third row, three examples represent primitives of ``subject touches object'', ``subject and object moving away'' and ``subject and object are person''. And the top ranked relation we return is \emph{adult, throw, child}.}
\label{fig:retrieval}
\end{figure}

\textbf{Video relation query-by-primitive-examples.} 
In Figure~\ref{fig:retrieval} we show three search cases, where for each case three primitive examples are given as input. We use the VidOR validation set for the search. The results show that we can find relevant video relations in space and time across many videos, simply by providing a few primitive examples, further highlighting the importance of compositions for video relations.

\section{Conclusion}
We propose an approach to video relation classification and detection that operates on pairs of object tubelets from the start. By doing so we no longer have to scatter the video into snippets or individual object tubelets and gather them at the end. To represent all pairs of object tubelets appearing in a video, we propose Social Fabric: an encoding built on a composition of data-driven interaction primitives, akin to the classical codebook approach. We use the encoding in a two-stage network, that first suggest proposals that are likely interacting and then fine-tunes and predicts it most likely predicate label. Experiments demonstrate the benefit of early video relation modeling, our encoding, as well as the two-stage architecture, leading to a new state-of-the-art on two video relation benchmarks. We also show how the encoding enables spatio-temporal video search by query-by-primitive-examples.

\textbf{Acknowledgements.} The authors thank Pengwan Yang for his help on figure design and comments.

{\small
\bibliographystyle{ieee_fullname}
\bibliography{social-fabric}
}

\end{document}